\documentclass[11pt,a4paper]{article}
\usepackage[hyperref]{acl2021}
\usepackage{times}
\usepackage{latexsym}
\usepackage{graphicx}
\usepackage{caption}
\usepackage{subcaption}
\usepackage{booktabs}

\usepackage{algpseudocode,algorithm,algorithmicx}
\newcommand*\Let[2]{\State #1 $\gets$ #2}
\usepackage{microtype}

\usepackage{microtype}

\aclfinalcopy 

\title{XeroAlign: Zero-Shot Cross-lingual Transformer Alignment}

\author{Milan Gritta \\
	Huawei Noah's Ark Lab, London \\
	\texttt{milan.gritta@huawei.com} \\ \And
	Ignacio Iacobacci  \\
	Huawei Noah's Ark Lab, London \\
	\texttt{ignacio.iacobacci@huawei.com} \\}

\date{}

\begin{document}
\maketitle
\begin{abstract}
	The introduction of pretrained cross-lingual language models brought decisive improvements to multilingual NLP tasks. However, the lack of labelled task data necessitates a variety of methods aiming to close the gap to high-resource languages. Zero-shot methods in particular, often use translated task data as a training signal to bridge the performance gap between the source and target language(s). We introduce \textbf{XeroAlign}, a simple method for task-specific alignment of cross-lingual pretrained transformers such as XLM-R. XeroAlign uses translated task data to encourage the model to generate similar sentence embeddings for different languages. The XeroAligned XLM-R, called XLM-RA, shows strong improvements over the baseline models to achieve state-of-the-art zero-shot results on three multilingual natural language understanding tasks. XLM-RA's text classification accuracy exceeds that of XLM-R trained with labelled data and performs on par with state-of-the-art models on a cross-lingual adversarial paraphrasing task.
\end{abstract}

\section{Introduction}

\begin{figure}[t]
	\centering
	\includegraphics[width=0.95\linewidth]{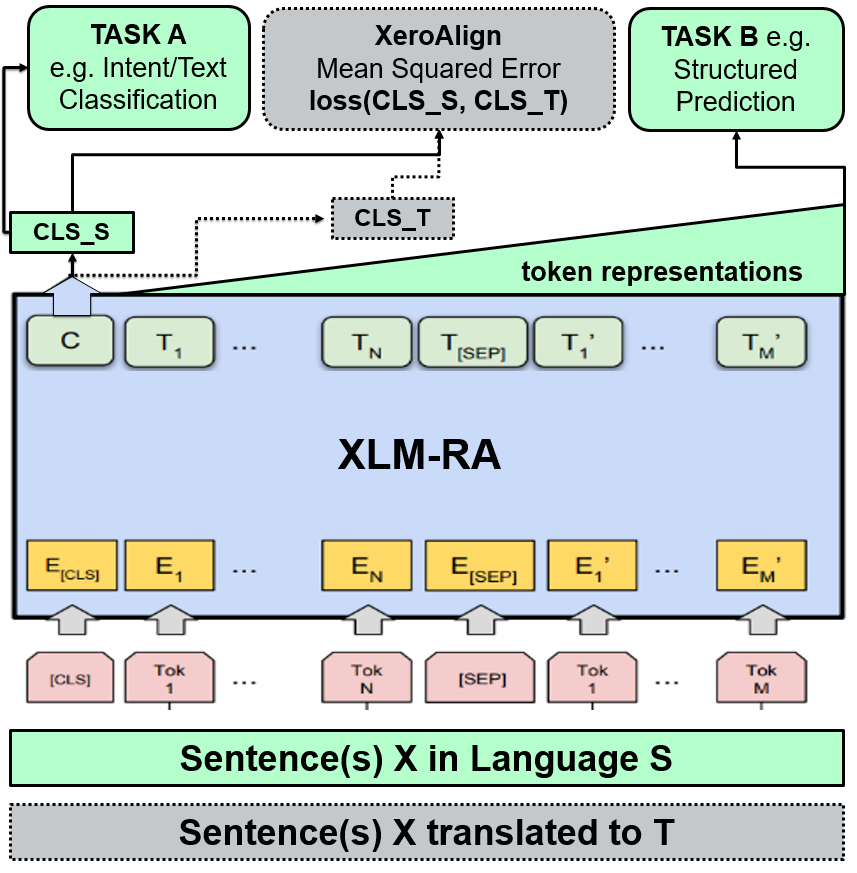}
	\caption{The XeroAligned XLM-R model (called XLM-RA) for cross-lingual NLU. The XeroAlign loss is added to the otherwise unaltered training to encourage the sentence embeddings in different languages to be similar, enabling zero-shot reuse of the classifier(s).}
	\label{fig:xlmra}
\end{figure}

In just a few years, transformer-based \cite{vaswani2017attention} pretrained language models have achieved state-of-the-art (SOTA) performance on many NLP tasks \cite{wang2019superglue}. Transfer learning enabled the self-supervised pretraining on unlabelled datasets to learn linguistic features such as syntax and semantics in order to improve tasks with limited training data \cite{wang2019glue}. Pretrained cross-lingual language models (PXLMs) have soon followed to learn general linguistic features and properties of dozens of languages \cite{lample2019cross,xue2020mt5}. For multilingual tasks, however, adequate labelled data is usually only available for a few well-resourced languages such as English. Zero-shot approaches were introduced to transfer the task knowledge to languages without the requisite training data. To this end, we introduce \textbf{XeroAlign}, a conceptually simple, efficient and effective method for task-specific alignment of sentence embeddings generated by PXLMs, aimed at effective zero-shot cross-lingual transfer. XeroAlign is an auxiliary loss function, which uses translated data (typically from English) to bring the zero-shot performance in the target language closer to the source (labelled) language, as illustrated in Figure \ref{fig:xlmra}. We apply our proposed method to the publicly available XLM-R transformer \cite{conneau2019unsupervised} but instead of pursuing large-scale model alignment with general parallel corpora such as Europarl \cite{koehn2005europarl}, we show that a simplified, task-specific model alignment is an effective and efficient approach to zero-shot transfer for cross-lingual natural language understanding (XNLU). We evaluate our method on 4 datasets that cover 11 unique languages. The XeroAligned XLM-R model (XLM-RA) achieves SOTA scores on three XNLU datasets, exceeds the text classification performance of XLM-R trained with labelled data and performs on par with SOTA models on an adversarial paraphrasing task.

\section{Related Work}
\label{related}

In order to cluster prior work, we formulate an approximate taxonomy in Table \ref{tab:related} for the purposes of positioning our approach in the most appropriate context. The relevant zero-shot transfer methods can generally be grouped by a) whether the alignment is targeted at each task, i.e. is task-specific [TS] or is task-agnostic [TA] and b) whether the alignment is applied to the model [MA] or data [DA]. Our contribution falls mostly into the [MA,TS] category although close methodological similarities are also found in the [MA,TA] group.

\begin{table}[h]
	\centering
	\begin{tabular}{@{}c|c|c@{}}\toprule
		Groups & \textit{Task-Specific} & \textit{Task-Agnostic} \\ \midrule
		\textit{Data Align} & [DA,TS] & No relevant work \\
		\textit{Model Align} & [MA,TS] & [MA,TA] \\ \bottomrule
	\end{tabular}
	\caption{An approximate taxonomy of prior work.}
	\label{tab:related}
\end{table}

\paragraph{Transformer-based PXLMs}

For transformer-based PXLMs, two basic types of representations are commonly used: 1) A sentence embedding for tasks such as text classification \cite{conneau2018xnli} or sentence retrieval \cite{zweigenbaum2018overview}, which use the \verb|[CLS]| representation of the full input sequence, and 2) Token embeddings, which are used for structured prediction \cite{pan2017cross} or Q\&A \cite{lewis2019mlqa}, requiring each token's contextualised representation for a per-token inference. While our method uses the \verb|[CLS]| embedding, other approaches based on Contrastive Learning have used both types of representations to obtain a sentence embedding.

\paragraph{Contrastive Pretraining}
\label{contrastive_learning}

The closest prior works are related to Contrastive Learning \cite{becker1992self} (CL). CL is a self-supervised framework designed to improve visual representations. Recent examples include Momentum Contrast (MoCo) \cite{he2020momentum} and SimCLR \cite{chen2020simple}, both of which achieved strong improvements on image classification. The essence of CL is to generate representations that are similar for positive examples and dissimilar for negative examples. CL-based methods in cross-lingual NLP replace negative samples, formerly augmented images, with random sentences in the target language, typically thousands of sentences. Positive examples comprise sentences translated into the target language. While CL may be applicable to large-scale, task-agnostic model alignment, large batches of negative samples are infeasible for small labelled datasets. Negative samples drawn randomly from a small dataset are likely related (possibly duplicates), which is why our proposed alignment uses only positive samples. The following contrastive alignments are task-agnostic methods aiming to improve generic cross-lingual representations with large parallel datasets. In contrast, we align the PXLM with translated task data, making our approach simpler and more efficient while showing a strong zero-shot transfer on each task.

[MA,TA] \citet{hu2020explicit} have proposed two objectives for cross-lingual zero-shot transfer a) sentence alignment and b) word alignment. While CL is not mentioned, the proposed sentence alignment closely resembles contrastive learning with one encoder (e.g. SimCLR). Taking the average of the contextualised token representations as the input representation (as an alternative to the \verb|[CLS]| token), the model predicts the correct translation of the sentence within a batch of negative samples. An improvement is observed for text classification tasks and sentence retrieval but not structured prediction. The alignment was applied to a 12-layer multilingual BERT and the scores are comparable to the translate-train baseline (translate data and train normally). Instead, we use one of the best publicly available models, XLM-R from Huggingface, as our starting point since an improvement in a weaker baseline is not guaranteed to work in a stronger model that may have already subsumed those upgrades during pretraining.

Contrastive alignment based on MoCo with two PXLM encoders was proposed by \citet{pan2020multilingual}. Using an L2 normalised \verb|[CLS]| token with a non-linear projection as the input representation, the model was aligned on 250K to 2M parallel sentences with added Translation Language Modelling (TLM) and a code-switching augmentation. No ablation for MoCo was provided to estimate its effect although the combination of all methods did provide improvements with multilingual BERT as the base learner. Another model inspired by CL is InfoXLM \cite{chi2020infoxlm}. InfoXLM is pretrained with TLM, multilingual Masked Language Modelling (mMLM) and Cross-lingual Contrastive Learning called XLCo. Like MoCo, they use two encoders that use the \verb|[CLS]| token (or the layer average) as the sentence representation, taken from layers 8 (base model) and 12 (large model). Ablation showed a 0.2-0.3 improvement in accuracy for XNLI and MLQA \cite{lewis2019mlqa}. Reminiscent of earlier work \cite{hermann2014multilingual}, the task-agnostic sentence embedding model \cite{feng2020language} called LaBSe (Language-agnostic BERT sentence embeddings) uses the \verb|[CLS]| representations of two BERT encoders (compared to our single encoder) with a margin loss and 6 billion parallel sentences to generate multilingual representations. While similarities exist, our multi-task alignment is an independently devised, more efficient, task-specific and a simplified version of the aforementioned approaches.

[DA,TS] Zero-shot cross-lingual models often use machine translation to provide a training signal. This is a straightforward data transformation for text classification tasks given that adequate machine translation models exist for many language pairs. However, for structured prediction tasks such as Slot Filling or Named Entity Recognition, the non-trivial task of \textit{aligning token/data labels} can lead to an improved cross-lingual transfer as well. One of the most used word alignment methods is fastalign \cite{dyer2013simple}. Frequently used as a baseline, it involves aligning the word indices in parallel sentences in an unsupervised manner, prior to regular supervised learning. In some scenarios, fastalign can approach SOTA scores for slot filling \cite{schuster2018cross}, however, the quality of alignment varies between languages and can even degrade performance \cite{li2020mtop} below baseline. An alternative data alignment approach called CoSDA \cite{qin2020cosda} uses code-switching as data augmentation. Random words in the input are translated and replaced to make model training highly multilingual, leading to improved cross-lingual transfer. Attempts were also made to automatically learn how to code-switch \cite{liu2020attention}. While improvements were reported, it's uncertain how much SOTA models would benefit.

[MA,TS] Continuing with label alignment for slot filling, \citet{xu2020end} tried to predict and align slot labels jointly during training instead of modifying data labels explicitly before fine-tuning. While soft-align improves on fastalign, the difficulty of label alignment makes it challenging to improve on the SOTA. For text classification tasks such as Cross-lingual Natural Language Inference \cite{conneau2018xnli}, an adversarial cross-lingual alignment was proposed by \citet{qi2020translation}. Adding a self-attention layer on top of multilingual BERT \cite{devlin2018bert} or XLM \cite{lample2019cross}, the model learns the XNLI task while trying to fool the language discriminator in order to produce language-agnostic input representations. While improvements over baselines were reported, the best scores were around 2-3 points behind the standard XLM-R model. 

\section{Methodology}

We introduce \textbf{XeroAlign}, a conceptually simple, efficient and effective method for task-specific alignment of sentence embeddings generated by PXLMs, aimed at effective zero-shot cross-lingual transfer. XeroAlign is an auxiliary loss function that is jointly optimised with the primary task, e.g. text classification and/or slot filling, as shown in Figure \ref{fig:xlmra}. We use standard architecture for each task and only add the minimum required number of new parameters. For text classification tasks, we use the \verb|[CLS]| token of the PXLM as our pooled sentence representation. A linear classifier (hidden size \verb|x| number of classes) is learnt on top of the \verb|[CLS]| embedding using cross-entropy as the loss function (TASK A in Figure \ref{fig:xlmra}). For slot filling, we use the contextualised representations of each token in the input sequence. Once again, a linear classifier (hidden size \verb|x| number of slots) is learnt with a cross-entropy loss (TASK B in Figure \ref{fig:xlmra}).\\

Algorithm \ref{alg:xlm-ra} shows a standard training routine augmented with XeroAlign. Let $\mathit{PXLM}$ be a pretrained cross-lingual transformer language model, $X$ be the standard English training data and $U$ be the machine translated parallel utterances (from $X$). Those English utterances were translated into each target language using our internal machine translation service. A public online translator e.g. Google Translate can also be used. For the PAWS-X task, we use the public version of the translated data\footnote{\url{https://github.com/google-research-datasets/paws}}. We then obtain the $CLS_S$ and $CLS_T$ embeddings by taking the first token of the $\mathit{PXLM}$ output sequence for the source $x_s$ and target $x_t$ sentences respectively. Using a Mean Squared Error loss function as our similarity function $sim$, we compute the distance/loss between $CLS_S$ and $CLS_T$. The sum of the losses ($total\_loss$) is then backpropagated normally. We have conducted all XeroAlign training as multi-task learning for the following reason. When the $\mathit{PXLM}$ is aligned first, followed by primary task training, the $\mathit{PXLM}$ exhibits poor zero-shot performance. Similarly, learning the primary task first, followed by XeroAlign fails as the primary task is partially unlearned during alignment. This is most likely due to the catastrophic forgetting problem in deep learning \cite{goodfellow2013empirical} hence the need for joint optimisation.

\begin{algorithm}[t]
	\begin{algorithmic}[1]
		\Let{$\mathit{PXLM}$}{Pretrained Cross-lingual LM}
		\Let{$X$}{Training data tuples in English}
		\Let{$U$}{Utterances translated into Target Lang.}
		\Let{$sim$}{similarity function e.g. MSE}
		\Statex
		\State \textit{\# training loop}
		\For{$(x_s, y), x_t \in X, U$}
		\Let{$task\_loss$}{$task\_loss\_fn(x_s, y)$}
		\Let{$CLS_S$}{$\mathit{PXLM}(x_s)$}
		\Let{$CLS_T$}{$\mathit{PXLM}(x_t)$}
		\Let{$align\_loss$}{$sim(CLS_S, CLS_T)$}
		\Let{$total\_loss$}{$task\_loss+align\_loss$}
		\State \textit{\# update model parameters}
		\EndFor
	\end{algorithmic}
	\caption{The XeroAlign algorithm. \label{alg:xlm-ra}}
\end{algorithm}

\subsection{Experimental Setup}

In order to make our method easily accessible and reproducible\footnote{Email \textbf{Milan Gritta} to request code and/or data.}, we use the publicly available XLM-R transformer from Huggingface \cite{wolf2019huggingface} built on top of PyTorch \cite{NEURIPS2019_9015}. We set a single seed for all experiments and a single learning rate for each dataset. No hyperparameter sweep was conducted to ensure a robust, low-resource, real-world deployment and to make a fair comparison with SOTA models. XLM-R was XeroAligned over 10 epochs and optimised using Adam \cite{kingma2014adam} and a OneCycleLR \cite{smith2019super} scheduler.

\subsection{Datasets}

We evaluate XeroAlign with four datasets covering 11 unique languages (en, de, es, fr, th, hi, ja, ko, zh, tr, pt) across three tasks (intent classification, slot filling, paraphrase detection).

\paragraph{PAWS-X} \cite{yang2019paws} is a multilingual version of PAWS \cite{zhang2019paws}, a binary classification task for identifying paraphrases. Examples were sourced from Quora Question Pairs\footnote{\url{https://www.quora.com/q/quoradata/First-Quora-Dataset-Release-Question-Pairs}} and Wikipedia, chosen to mislead simple `word overlap' models. PAWS-X contains 4,000 random examples from PAWS, for the development and test set, covering seven languages (en, de, es, fr, ja, ko, zh), totalling 48,000 human translated paraphrases. We use the multilingual train sets that contain approximately 49K machine translated examples.

\paragraph{MTOD} is a Multilingual Task-Oriented Dataset provided by \citet{schuster2018cross}. It covers three domains (alarm, weather, reminder) and three languages of different sizes: English (43K), human-translated Spanish (8.3K) and Thai (5K). MTOD comprises two correlated NLU tasks, intent classification and slot filling. The SOTA scores are reported by \citet{li2020mtop} and \citet{schuster2018cross}.

\paragraph{MTOP} is a Multilingual Task-Oriented Parsing dataset provided by \citet{li2020mtop} that covers interactions with a personal assistant. We use the standard flat version, which has the highest reported zero-shot SOTA scores by \citet{li2020mtop}. A tree-like compositional version of the data designed for nested queries is also provided. MTOP contains 100K+ human-translated examples in 6 languages (en, de, es, fr, th, hi) spanning 11 domains. 

\paragraph{MultiATIS++} by \citet{xu2020end} is an extension of the Multilingual version of ATIS \cite{upadhyay2018almost}, initially translated into Hindi and Turkish only. Six new (human-translated\footnote{We have encountered some minor issues with slot annotations. Around 60-70 entities across 5 languages (fr, zh, hi, ja, pt) had to be corrected as the number of slot tags did not agree with the number of tokens in the sentence. However, this only concerns a tiny fraction of the $\sim$400k+ tags/tokens covered by those languages. We are happy to share the corrections, too.}) languages (de, es, fr, zh, ja, pt) were added with $\sim$4 times as many examples each (around 6K per language) for 9 languages in total. Both of these datasets are based on the original English-only ATIS \cite{price1990evaluation} featuring users interacting with an automated air travel information service (via intent recognition and slot filling tasks).

\begin{table*}[t]
	\centering
	\setlength{\tabcolsep}{7pt}
	\begin{tabular}{l|ccccc|c}
		\toprule \textbf{Model} & \textbf{Spanish} & \textbf{French} & \textbf{German} & \textbf{Hindi} & \textbf{Thai} & \textbf{Average} \\ \midrule
		XLM-R Target & 95.9 / 91.2 & 95.5 / 89.6 & 96.6 / 88.3 & 95.1 / 89.1 & 94.8 / 87.7 & 95.6 / 89.2 \\ \midrule
		XLM-R 0-shot & 91.9 / 84.3 & 93.0 / \textbf{83.7} & 87.5 / 80.7 & 91.4 / 76.5 & 87.6 / 55.6 & 90.3 / 76.2 \\
		XLM-RA & \textbf{96.6} / 84.4 & \textbf{96.5} / 83.3 & \textbf{95.7} / \textbf{84.5} & \textbf{95.2} / \textbf{80.1} & \textbf{94.1} / \textbf{69.1} & \textbf{95.6} / \textbf{80.3} \\ 
		\citet{li2020mtop} & 96.3 / \textbf{84.8} & 95.1 / 82.5 & 94.8 / 83.1 & 94.2 / 76.5 & 92.1 / 65.6 & 94.5 / 77.9 \\
		\bottomrule
	\end{tabular}
	\caption{MTOP results as Intent Classification Accuracy / Slot Filling F-Score. Best English scores: 97.3 / 93.9.}
	\label{tab:mtop_detailed}
\end{table*}

\subsection{Metrics}

We use standard evaluation metrics, that is, accuracy for paraphrase detection and intent classification, F-Score\footnote{\url{https://pypi.org/project/seqeval/}} for slot filling.

\section{Results and Analysis}

We use `XLM-R Target' to refer to model performance on the labelled target language. We provide zero-shot scores (denoted `XLM-R 0-shot'), the XLM-RA results and the reported SOTA figures. For PAWS-X, we provide a second baseline called `Translate-Train', which comprises the union of Target and English train data. Scores are given for the large\footnote{Large=24 layers, 550M par, Base=12 layers, 270M par.} model unless specified otherwise. \\

The XeroAligned XLM-R achieves state-of-the-art scores on three task-oriented XNLU datasets. For MTOP (Table \ref{tab:mtop_detailed}), the intent classification accuracy (+1.1) and slot filling F-Score (+2.4) averaged over 5 languages improved on XLM-R-Large with translated utterances, slot label projection and distant supervision \cite{li2020mtop}. For MultiATIS++ (Table \ref{tab:multi_atis}), XLM-RA shows an improved intent accuracy (+1.1) and slot F-Score (+3.2) over 8 languages, as compared to a large multilingual BERT with translated utterances and slot label softalign \cite{xu2020end}. For MTOD (Table \ref{tab:mtod}), the classification accuracy (+1.3) and slot tagging F-Score (+5.0) on average improved on XLM-R-Large with translated utterances, slot label projection and distant supervision \cite{li2020mtop}. MTOD is the only dataset where the XLM-RA-base model outperforms (albeit marginally) XLM-RA-large. Finally, we also compare our intent classification accuracy (+8.1) and slot filling F-Score (+8.7) for the MTOD dataset to a BiLSTM with translated utterances and slot label projection \cite{schuster2018cross}, which had the SOTA F-Score for Thai.\\

On the adversarial paraphrase task (PAWS-X, Table \ref{tab:paws_x}), averaged over 7 languages, XLM-RA scores marginally higher (+0.1 accuracy) than VECO \cite{luo2020veco}, a variable cross-lingual encoder-decoder and marginally lower (-0.2 accuracy) than FILTER \cite{fang2020filter}, an enhanced cross-lingual fusion model, which was the SOTA until 01/2021. We now turn our attention to the improvements over 'vanilla' zero-shot XLM-R.

\begin{table*}[t]
	\centering
	\setlength{\tabcolsep}{8pt}
	\begin{tabular}{l|ccccccc|c}
		\toprule \textbf{Model} & \textbf{EN} & \textbf{DE} & \textbf{ES} & \textbf{FR} & \textbf{JA} & \textbf{KO} & \textbf{ZH} & \textbf{Average} \\ \midrule
		XLM-R Target & 95.6 & 90.9 & 92.5 & 92.4 & 85.1 & 86.4 & 87.2 & 90.0 \\
		XLM-R Translate-Train & 95.7 & 91.6 & 92.3 & 92.5 & 85.2 & 85.8 & 87.7 & 90.1 \\ \midrule
		XLM-R 0-shot & 95.6 & 91.0 & 91.1 & 91.9 & 81.7 & 81.6 & 85.4 & 88.3 \\
		\citet{luo2020veco} & \textbf{96.4} & \textbf{93.0} & \textbf{93.0} & 93.5 & 87.2 & 86.8 & 87.9 & 91.1 \\
		XLM-RA & 95.8 & 92.9 & \textbf{93.0} & \textbf{93.9} & 87.1 & 87.1 & 88.9 & 91.2 \\
		\citet{fang2020filter} & 95.9 & 92.8 & \textbf{93.0} & 93.7 & \textbf{87.4} & 87.6 & \textbf{89.6} & \textbf{91.4} \\
		\toprule
		\multicolumn{9}{c}{\textit{Section \ref{domain-shift} experiment below: aligning with development/test set utterances but no task labels.}} \\ \midrule
		XLM-RA (Exp) & 95.8 & 94.2 & 94.4 & 94.8 & 91.6 & 92.6 & 92.1 & 93.6 \\ \bottomrule
	\end{tabular}
	\caption{PAWS-X results as Paraphrase Classification Accuracy. }
	\label{tab:paws_x}
\end{table*}

\subsection{Zero-shot Text Classification}

The intent classification accuracy of our XeroAligned XLM-R exceeds that of XLM-R trained with labelled data, averaged across three task-oriented XNLU datasets and 15 test sets (Tables \ref{tab:mtop_detailed}, \ref{tab:multi_atis} and \ref{tab:mtod}). Starting from an already competitive baseline model, XeroAlign improves intent classification by $\sim$5-10 points (larger for XLM-R-base, see Table \ref{tab:base} in Section \ref{smaller}). The benefits of cross-lingual alignment are particularly evident in low-resource languages (tr, hi, th), which is encouraging for real-world applications with limited resources. Zero-shot paraphrase detection is another instance of text classification. We report XLM-RA accuracy in Table \ref{tab:paws_x}, which exceeds both Target and the Translate-Train averages by over 1 point and by almost 3 points over the zero-shot XLM-R baseline (even mores for XLM-RA-base). \\

Note that the \textit{amount of training data} is the same for XeroAlign and Target (except MTOD) thus there is no advantage from using additional data. The primary task, which is learnt in English, has a somewhat higher average performance ($\sim$1.5 points) than the Target languages. We hypothesise that transferring this advantage from a high-resource language via \textbf{XeroAlign} is the primary reason behind its effectiveness compared to using target data directly. Given that Target performance has recently been exceeded with MoCo \cite{he2020momentum} and the similarities between contrastive learning and XeroAlign, our finding seems in line with recent work, which is subject to ongoing research \cite{zhao2020makes}.

\subsection{Zero-shot Structured Prediction}

While XLM-RA is able to exceed Target accuracy for text classification tasks, even our best F-Scores for slot filling are 8-19 points behind Target accuracy. This is despite a strong average improvement of +4.1 on MTOP, +5.7 on MultiATIS++ and +5.2 on MTOD for the XLM-R-large model (greater for the XLM-RA-base model). We think the gap is primarily down to the difficulty of the sequence labelling task, i.e. zero-shot text classification is `easier' than zero-shot slot filling, which is manifested by a $\sim$10-20 point gap between scores. Sentences in various languages have markedly different input lengths and token/entity order thus word-level inference in cross-lingual zero-shot settings becomes significantly more challenging than sentence-level prediction because syntax plays a less critical role in sequence classification.\\

A less significant reason, related to XeroAlign's architecture, may be our choice to align the PXLM on the \verb|[CLS]| embedding, which is subsequently used `as is' for text classification tasks. Aligning individual token representations through the \verb|[CLS]| embedding improves structured prediction as well, however, as the token embeddings are not directly used, the parameters in the uppermost transformer layer (following Multi-Head Attention) never receive any gradient updates from XeroAlign. Closing this gap is a challenging opportunity, which we reserve for future work. Once again, the languages with lower NLP resources (th, hi, tr) tend to benefit the most from cross-lingual alignment.

\subsection{XeroAlign Generalisation}

We briefly want to investigate the generalisation of XeroAlign, taking the PAWS-X task as our use case. We are interested in fining out whether aligning on \textit{just one language} has any zero-shot benefits for other languages. Table \ref{tab:one_language} shows the XLM-RA results when aligned on a single language (rows) and tested on other languages (columns).

\begin{table}[h]
	\centering
	\small
	\setlength{\tabcolsep}{4pt}
	\begin{tabular}{c|c|cccccc|c}
		\toprule - & \textbf{EN} & \textbf{DE} & \textbf{ES} & \textbf{FR} & \textbf{JA} & \textbf{KO} & \textbf{ZH} & \textbf{AVE} \\ \midrule
		\textbf{DE} & \textbf{96.0} & \textbf{92.9} & 92.3 & 92.6 & 84.0 & 84.5 & 86.5 & 89.8 \\
		\textbf{ES} & 95.9 & 92.6 & \textbf{93.0} & 93.1 & 83.9 & 84.1 & 86.4 & 89.9 \\
		\textbf{FR} & 95.9 & 92.5 & 92.9 & \textbf{93.9} & 83.9 & 84.1 & 86.9 & 90.0 \\
		\textbf{JA} & \textbf{96.0} & 92.6 & 91.8 & 93.1 & \textbf{87.1} & \textbf{87.4} & 87.9 & \textbf{90.8} \\
		\textbf{KO} & 95.7 & 92.6 & 92.0 & 92.7 & 80.6 & 87.1 & 87.3 & 90.5 \\
		\textbf{ZH} & 95.5 & 92.0 & 92.6 & 92.7 & 86.3 & 86.2 & \textbf{88.9} & 90.6 \\ \midrule
		\textbf{EU} & 96.2 & 92.5 & 93.0 & 94.1 & 84.9 & 85.2 & 87.1 & 90.4 \\
		\textbf{AS} & 96.0 & 93.0 & 92.1 & 92.7 & 85.9 & 87.6 & 88.4 & 90.8 \\\bottomrule
	\end{tabular}
	\caption{XLM-RA aligned on \textbf{one} PAWS-X language (rows), evaluated on others (columns). AVE = average. EU = European languages, AS = Asian languages.}
	\label{tab:one_language}
\end{table}

We can see that aligning on Asian languages (Japanese in particular) attains the best average improvement compared to aligning with European languages. This seems to reflect the known performance bias of XLM-R towards (high-resource) European languages, all of which show a strong improvement, regardless of language. Aligning only on European languages (de, es, fr) improves the average to 90.4 but aligning on Asian languages (zh, ko, ja) does not improve over Japanese (90.8). In any case, it is notable that the XLM-R model XeroAligned on \textit{just a single language} is able to carry this advantage well beyond a single language thus improve average accuracy by 1.5-2.5 points over baseline (88.3) from Table \ref{tab:paws_x}. This effect is even stronger for MTOP (+4 accuracy, +3 F-Score).

\begin{table*}[t]
	\centering
	\small
	\setlength{\tabcolsep}{3.7pt}
	\begin{tabular}{l|cccccccc|c}
		\toprule \textbf{Model} & \textbf{DE} & \textbf{ES} & \textbf{FR} & \textbf{TR} & \textbf{HI} & \textbf{ZH} & \textbf{PT} & \textbf{JA} & \textbf{AVE} \\ \midrule
		XLM-R Target & 97.0/95.3 & 97.3/87.9 & 97.8/93.8 & 80.6/74.0 & 89.7/84.1 & 95.5/95.9 & 97.2/94.1 & 95.5/92.6 & 93.8/89.7 \\ \midrule
		XLM-R 0-shot & 96.4/84.8 & 97.0/85.5 & 95.3/81.8 & 76.2/41.2 & 91.9/68.2 & 94.3/82.5 & 90.9/81.9 & 89.8/77.6 & 91.5/75.5 \\
		XLM-RA & \textbf{97.6}/84.9 & \textbf{97.8}/\textbf{85.9} & 95.4/\textbf{81.4} & 93.4/\textbf{70.6} & \textbf{94.0}/\textbf{79.7} & \textbf{96.4}/83.3 & \textbf{97.6}/79.9 & \textbf{96.1}/\textbf{83.5} & \textbf{96.0}/\textbf{81.2} \\
		\citet{jain2019entity} & 96.0/87.5 & 97.0/84.0 & 97.0/79.8 & \textbf{93.7}/44.8 & 92.4/77.2 & 95.2/\textbf{85.1} & 96.5/\textbf{81.7} & 88.5/82.6 & 94.5/77.8\\ 
		\citet{xu2020end} & 96.7/\textbf{89.0} & 97.2/76.4 & 97.5/79.6 & \textbf{93.7}/61.7 & 92.8/78.6 & 96.0/83.3 & 96.8/76.3 & 88.3/79.1 & 94.9/78.0\\ \bottomrule
	\end{tabular}
	\caption{MultiATIS++ as Intent Classification Accuracy / Slot Filling F1. English model: 97.9/97. AVE = average.}
	\label{tab:multi_atis}
\end{table*} 

\begin{table}[t]
	\centering
	\setlength{\tabcolsep}{4.3pt}
	\begin{tabular}{l|cc|c}
		\toprule 
		\textbf{Model} & \textbf{Spanish} & \textbf{Thai} & \textbf{AVE} \\ \midrule
		$\mathsection$ Target (B) & 98.7/89.1 & 96.8/93.1 & 97.8/91.1 \\
		$\mathsection$ Target (L) & 98.8/89.8 & 97.8/94.4 & 98.3/92.1 \\ \midrule
		$\mathsection$ 0-shot (B) & 90.7/70.1 & 71.9/53.1 & 81.3/61.6 \\
		$\mathsection$ 0-shot (L) & 97.1/85.7 & 82.8/47.7 & 90.0/66.7 \\
		XLM-RA (B) & 98.9/86.9 & 97.9/\textbf{60.2} & 98.4/\textbf{73.6} \\
		XLM-RA (L) & \textbf{99.2}/\textbf{88.4} & \textbf{98.4}/57.3 & \textbf{98.8}/72.9 \\
		\citeauthor{schuster2018cross} & 85.4/72.9 & 95.9/55.4 & 90.7/64.2 \\
		\citeauthor{li2020mtop} & 98.0/83.0 & 96.9/52.8 & 97.5/67.9 \\
		\bottomrule
	\end{tabular}
	\caption{MTOD results as Intent Classification Accuracy / Slot Filling F-Scores. Our best English score: 99.3/96.6, (B) = Base, (L) = Large, $\mathsection$ = XLM-R model.}
	\label{tab:mtod}
\end{table}

\subsection{Smaller Language Models}
\label{smaller}

We observed that the XeroAligned XLM-R-base model shows an even greater improvement than its larger counterpart with 24 layers and 550M parameters. To this end, we report the XLM-RA-base results (12 layers, 270M parameters) in Table \ref{tab:base} as the average scores over all languages for MTOP, PAWS-X, MTOD and MultiATIS++. We use a relative \% improvement over the baseline XLM-R to compare the models fairly. The paraphrase detection accuracy improves by 3.3\% for the large (L) PXLM versus 6.5\% for the base (B) model. \\

\begin{table}[h]
	\centering
	\small
	\setlength{\tabcolsep}{4.5pt}
	\begin{tabular}{l|cccc}
		\toprule 
		\textbf{Model} & \textbf{MTOP} & \textbf{PAWS-X} & \textbf{M-ATIS} & \textbf{MTOD}\\ \midrule
		$\mathsection$ Target & 94.0/88.1 & 85.2 & 89.0/86.3 & 97.6/92.2 \\ \midrule
		$\mathsection$ 0-shot & 80.8/68.9 & 81.7 & 76.9/65.0 & 80.1/64.8\\
		XLM-RA & 93.3/78.9 & 87.0 & 93.0/73.4 & 98.5/74.7\\ \bottomrule
	\end{tabular}
	\caption{The XLM-R(A) base model averages as intent classification accuracy / slot filling F-Score (or paraphrase accuracy for PAWS-X). $\mathsection$ = XLM-R model.}
	\label{tab:base}
\end{table}

Across three XNLU datasets, XeroAlign improves the standard XLM-R by 9.5\% (L) versus 14.2\% (B) on structured prediction (slot filling) and by 7.1\% (L) versus 19.8\% (B) on text classification (intent recognition). Therefore, applications with lower computational budgets can also achieve competitive performance with our simple cross-lingual alignment method for transformed-based PXLMs. In fact, the base XLM-RA can reach (on average) up to 90-95\% of the performance of its larger sibling using lower computational resources.

\subsection{Discussion}
\label{domain-shift}

The XLM-RA intent classification accuracy is (on average) within $\sim$1.5 points of English accuracy across three task-oriented XNLU datasets. However, the PAWS-X paraphrase detection accuracy is almost 5 points below English models, which is also the case for other state-of-the-art PXLMs in Table \ref{tab:paws_x}. Why does XLM-R struggle to generalise more on this task for languages other than English? We can exclude translation issues since all models used the publicly available PAWS-X machine-translated data. Instead, we think that the greater than expected deficit may be caused by a) domain/topic shift within the dataset and b) a possible data leakage for English. The original PAWS data \cite{zhang2019paws} was sourced from Quora Question Pairs and Wikipedia with neither being limited to any particular domain. As the English Wikipedia provides a large chunk of the English training data for XLM-R, it is possible that some of the English PAWS sentences may have been seen in training, which could explain the smaller generalisation gap for English. \\

We also want to find out whether this gap will diminish if we artificially remove the domain shift. To this end, we use parallel utterances (but not task labels) from the development and test sets to XeroAlign the XLM-R on an extended vocabulary that may not be present in the train set. We observe that the (Exp) model in Table \ref{tab:paws_x} shows an average improvement of over 2 points compared to the best XLM-RA and other SOTA models suggesting that the increased generalisation gap may be caused by a domain shift for non-English languages on this task. When that topic shift gets (perhaps artificially) removed, the model is able to bring accuracy back within $\sim$2 points of the English model (in line with XNLU tasks). Note that this effect can be masked for English due to the language biases in data used for pretraining. \\

In section \ref{contrastive_learning}, we outlined the most conceptually similar methods that conducted large-scale model pretraining with task-agnostic parallel sentence alignment as part of the training routine \cite{hu2020explicit,feng2020language,pan2020multilingual,chi2020infoxlm}. Where ablation studies were provided, the average improvement attributed to contrastive alignment was $\sim$0.2-0.3 points (though the tasks were slightly different). While we do not directly compare XeroAlign to contrastive alignment, it seems that task-specific alignment may be a more effective and efficient technique to improve zero-shot transfer, given the magnitude of our results. This leads us to conclude that the effectiveness of our method comes primarily from cross-lingual alignment of the task-specific vocabulary. Language is inherently ambiguous, the semantics of words and phrases shift somewhat from topic to topic, therefore, a cross-lingual alignment of sentence embeddings \textit{within the context of the target task} should lead to better results. Our simplified, lightweight method only uses translated task utterances, a single encoder model and positive samples, the alignment of which is challenging enough without arbitrary negative samples. In fact, this is the main barrier for applying contrastive alignment in task-specific NLP scenarios, i.e. the lack of carefully constructed negative samples. For smaller datasets, random negative samples would mean that the task is either too easy to solve, resulting in no meaningful learning or the model would receive conflicting signals by training on false positive examples, leading to degenerate learning. 

\subsection{Future Work}

Our recommendations for avenues of promising follow-up research involve any of the following: i) aligning more tasks such as Q\&A, Natural Language Inference, Sentence Retrieval, etc. ii) including additional languages, especially low-resource ones \cite{joshi2020state} and iii) attempting large-scale, task-agnostic alignment of PXLMs followed by task-specific alignment, which is reminiscent of the common transfer learning paradigm of pretraining with Masked Language Modelling before fine-tuning on the target task. To that end, there is already some emergent work on monolingual fine-tuning with an additional contrastive loss \cite{gunel2020supervised}. For the purposes of multilingual benchmarks \cite{hu2020xtreme,Liang2020XGLUEAN} or other pure empirical pursuits, an architecture or a language-specific hyperparameter search should optimise XLM-RA for significantly higher performance as the large transformer does not always outperform its smaller counterpart and because our hyperparameters remained fixed for all languages. Most importantly, the follow-up work needs to improve zero-shot transfer for cross-lingual \textit{structured prediction} such as Named Entity Recognition \cite{pan2017cross}, POS Tagging \cite{nivre2016universal} or Slot Filling \cite{schuster2018cross}, which is still lagging behind Target scores.

\section{Conclusions}

We have introduced \textbf{XeroAlign}, a conceptually simple, efficient and effective method for task-specific alignment of sentence embeddings generated by PXLMs, aimed at effective zero-shot cross-lingual transfer. XeroAlign is an auxiliary loss function that is easily integrated into the unaltered primary task/model. XeroAlign leverages translated data to bring the sentence embeddings in different languages closer together. We evaluated XeroAligned XLM-R models (named XLM-RA) on zero-shot cross-lingual text classification, adversarial paraphrase detection and slot filling tasks, achieving SOTA (or near-SOTA) scores across 4 datasets covering 11 unique languages. Our ultimate vision is a level of zero-shot performance at or near that of Target. The XeroAligned XLM-R partially achieved that goal by exceeding the intent classification and paraphrase detection accuracies of XLM-R trained with labelled data.



\bibliography{acl2021}
\bibliographystyle{acl_natbib}



	
\end{document}